\documentclass[letterpaper, 10 pt, conference]{ieeeconf}  

\IEEEoverridecommandlockouts                              

\usepackage{amsmath} 
\usepackage{amssymb}  
\usepackage{graphicx}
\usepackage{booktabs} 
\usepackage{multirow} 
\usepackage{xcolor} 
\usepackage{caption}
\usepackage{cite}
\makeatletter
\let\NAT@parse\undefined
\makeatother
\usepackage{hyperref}
\hypersetup{
    colorlinks=true,
    urlcolor=darkblue,
    linkcolor=darkgreen,
    citecolor=darkgreen
    }

\definecolor{darkgreen}{rgb}{0, 0.3, 0.6}
\definecolor{darkblue}{rgb}{0, 0.4, 0.6}
\definecolor{darkred}{rgb}{0.9, 0.2, 0}

\newcommand{\cmark}{\checkmark}
\newcommand{\xmark}{\times}

\title{\LARGE \bf
RoboEye: Enhancing 2D Robotic Object Identification with 

Selective 3D Geometric Keypoint Matching
}

\author{Xingwu Zhang$^{1}$, Guanxuan Li$^{1}$, Zhuocheng Zhang$^{1}$ and Zijun Long$^{1}$$^{\ast}$ 
\thanks{$^{1}$Hunan University}%
\thanks{*Corresponding author {\tt\small longzijun@hnu.edu.cn}}
}

\begin{document}

\maketitle
\thispagestyle{empty}
\pagestyle{empty}

\begin{abstract}
The rapidly growing number of product categories in large-scale e-commerce makes accurate object identification for automated packing in warehouses substantially more difficult. As the catalog grows, intra-class variability and a long tail of rare or visually similar items increase, and—when combined with diverse packaging, cluttered containers, frequent occlusion, and large viewpoint changes—these factors amplify discrepancies between query and reference images, causing sharp performance drops for methods that rely solely on 2D appearance features. Thus, we propose RoboEye, a two-stage identification framework that dynamically augments 2D semantic features with domain-adapted 3D reasoning and lightweight adapters to bridge training–deployment gaps. In the first stage, we train a large vision model to extract 2D features for generating candidate rankings. A lightweight 3D-feature-awareness module then estimates 3D feature quality and predicts whether 3D re-ranking is necessary, preventing performance degradation and avoiding unnecessary computation. When invoked, the second stage uses our robot 3D retrieval transformer, comprising a 3D feature extractor that produces geometry-aware dense features and a keypoint-based matcher that computes keypoint-correspondence confidences between query and reference images instead of conventional cosine-similarity scoring. Experiments show that \emph{RoboEye} improves Recall@1 by 7.1\% over the prior state of the art (RoboLLM). Moreover, \emph{RoboEye} operates using only RGB images, avoiding reliance on explicit 3D inputs and reducing deployment costs.  The code used in this paper is publicly available at: \url{https://github.com/longkukuhi/RoboEye}.

\end{abstract}

\section{INTRODUCTION}

\begin{figure}
    \centering
    \includegraphics[width=1.0\linewidth]{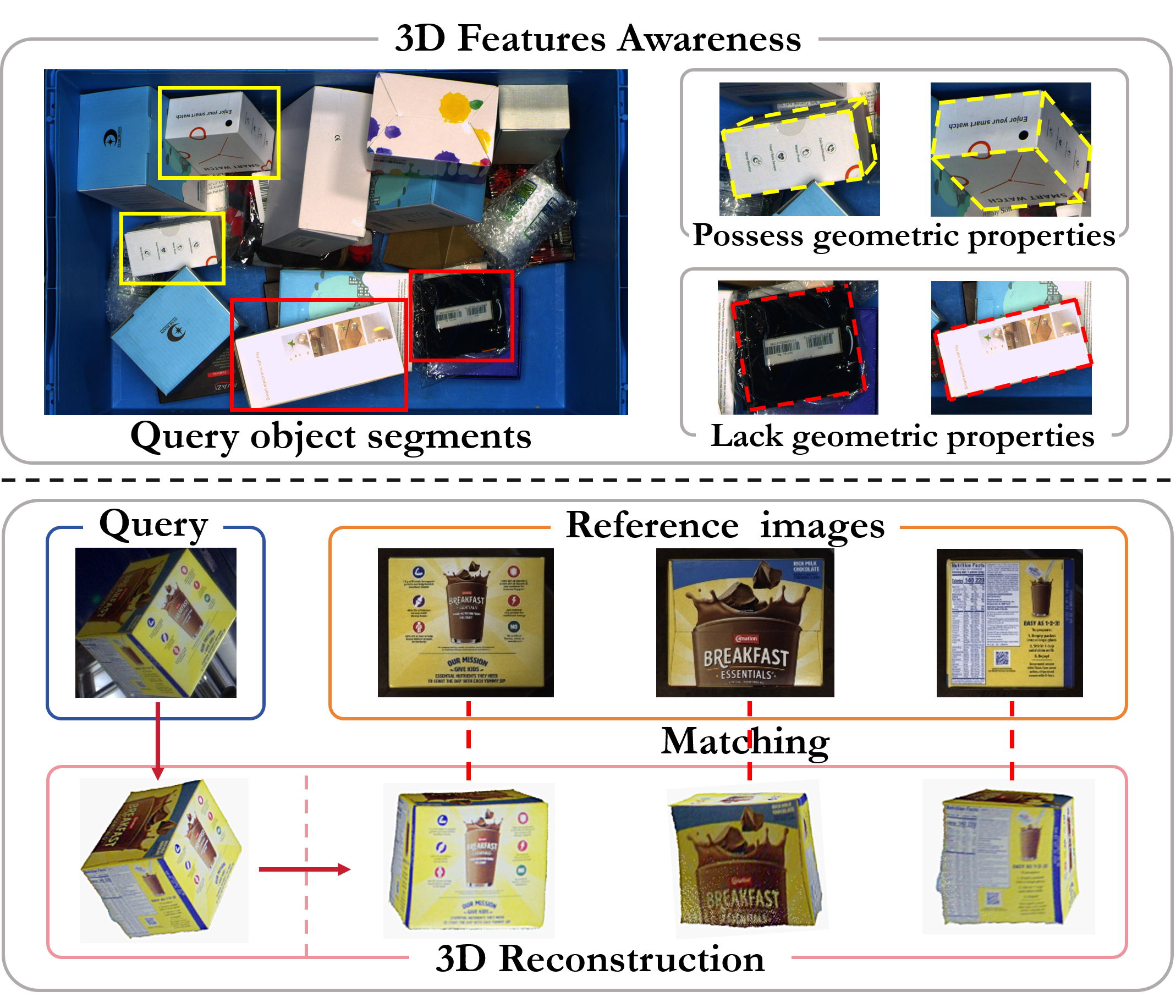}
    \caption{Object identification under warehouse conditions. Upper part: a cluttered container with candidate items, where some query segments provide sufficient geometric cues for 3D reasoning while others lack reliable geometry. Lower part: query–reference discrepancies from viewpoint and packaging variations, where 3D reconstruction enables more robust verification.}
    \label{fig:demo}
    \vspace{-7mm}
\end{figure}

Object identification (ID) aims to identify the category of a query image, which is fundamental to warehouse automation. In the pre-pick stage, correct ID within the source container provides semantic and geometric attributes (e.g., material properties and grasp points) required for motion planning~\cite{tang2025embodiment} and control~\cite{noh2025graspsam}; in the post-pick stage, ID governs handling and order-fulfillment accuracy, and at the e-commerce scale even small misidentification rates can produce substantial financial losses~\cite{aduh2024avoiding}. As reported in \cite{news}, Amazon’s Q1 2025 included about US\$1 billion in charges related to ``customer returns and tariff-induced inventory adjustments."

The rapid expansion of product catalogs, coupled with the complexity of robotic warehouse environments, makes accurate ID for automated packing increasingly challenging~\cite{correll2016analysis,mitash2023armbench}. Effective image identification requires visual features that are simultaneously discriminative and robust to operational variations. The objective is to maximize similarity between query and reference images of the same class (yielding high Recall@k) while minimizing similarity across different classes. Four major factors undermine this objective: viewpoint and pose variation, occlusion and clutter, packaging and intra-class variability, and inter-class similarity combined with long-tail distributions. As catalogs scale, these factors interact to introduce more visually similar items and sparser viewpoint coverage, crowding the feature space and reducing separability.

The ARMBench benchmark from Amazon~\cite{mitash2023armbench} exemplifies these conditions, showing that state-of-the-art systems (e.g., RoboLLM~\cite{long2024robollm}) relying exclusively on 2D appearance cues—such as texture, color, and local gradients—are particularly vulnerable. Because these cues lack invariance to viewpoint shifts, occlusions, and packaging variations, their discriminative power sharply deteriorates under real-world warehouse settings (shown in Figure~\ref{fig:demo})~\cite{eppner2016lessons,back2025high}. Consequently, methods based solely on 2D features fail to generalize reliably across these challenging scenarios.

To address these challenges, we propose leveraging 3D geometric features, which provide viewpoint-invariant characteristics and are inherently less sensitive to the problematic variations encountered in warehouse environments. A straightforward approach would be to incorporate explicit 3D inputs to complement 2D features and mitigate viewpoint-induced discrepancies. However, explicit 3D data (e.g., point clouds or depth maps) requires specialized sensors such as LiDAR or depth cameras, which increase deployment costs and complicate large-scale integration. This leads to the central question of this work:
\begin{quote}
\emph{How can 3D geometric cues be used to improve ID robustness under challenging warehouse conditions, without relying on explicit 3D inputs?}
\end{quote}
To answer this, we introduce \textbf{RoboEye}, a two-stage identification framework that augments strong 2D representations with selective, domain-adapted implicit 3D reasoning and lightweight adapters to bridge training–deployment gaps. This hybrid design both reinforces appearance-based signals and supplies viewpoint-invariant geometric cues, enabling robust identification when 2D features alone are insufficient. Specifically, in the first stage (see the upper part of Figure~\ref{fig:placeholder}), a large vision model produces robust 2D embeddings for initial ranking. A lightweight \emph{3D-feature-awareness} module with an MRR-driven 3D-awareness training scheme then determines whether geometric cues in the input image can be effectively exploited (demonstrated in the upper part of Figure~\ref{fig:demo}), avoiding unnecessary 3D computation when 2D features are already discriminative and preventing performance degradation from noisy 3D cues. If 3D cues are deemed useful, the second stage invokes our proposed robot 3D retrieval transformer, which integrates a multi-view 3D feature extractor with a keypoint-based retrieval matcher. The extractor generates geometry-aware representations across views, while the matcher evaluates keypoint correspondences between query and reference images.

Our main contributions are:
\begin{itemize}
\item We propose \textit{RoboEye}, the first framework to dynamically augment 2D appearance-based retrieval with domain-adapted implicit 3D geometric re-ranking, enabling robust identification without explicit 3D inputs.
\item We develop an MRR-driven 3D-awareness training scheme to train the \emph{3D-feature-awareness} module that selectively activates 3D re-ranking only when beneficial.
\item We introduce a 3D keypoint-based retrieval matcher that establishes confidence-weighted keypoint correspondences, offering a more robust similarity measure than conventional cosine similarity.
\item We develop an adapter-based training strategy for the robot 3D retrieval transformer, enabling efficient domain adaptation.
\item Our RoboEye outperforms the previous state of the art, RoboLLM, by up to 7.1\% on Recall@1, as demonstrated by extensive experiments on the Amazon ARMBench dataset.
\end{itemize}

\section{RELATED WORK}

\subsection{Robotic Object Identification}


The Object Identification (ID) task in ARMBench is essentially a specialized image retrieval problem~\cite{mitash2024scaling}, where a segmented query is matched against a reference gallery. This has long been studied in computer vision with applications in robotic manipulation~\cite{zeng2022robotic,di2024dinobot}, visual localization~\cite{anoosheh2019night}, and place recognition~\cite{chen2017only}. Early works such as DoUnseen~\cite{gouda2023dounseen} leveraged transformer-based classifiers with siamese matching, while later methods improved robustness via multimodal feature fusion~\cite{mitash2024scaling} and centroid triplet loss for variable-sized inputs~\cite{gouda2024learning}. Most recently, RoboLLM~\cite{long2024robollm} integrated the large vision model BEiT-3~\cite{wang2023image} with contrastive training, achieving state-of-the-art performance with lightweight adaptations. However, most existing approaches rely mainly on 2D features without fully exploiting the 3D geometric cues inherent in observations, which are crucial for robust ID under challenging warehouse conditions such as viewpoint shifts, occlusions, and packaging variations.
\vspace{-0.3mm}

\subsection{3D Geometric Features for Identification}

Comprehending 3D geometric features is imperative for practical applications such as robotics~\cite{yan2018learning,li2024representing} and autonomous driving~\cite{li2022important,sun2025sparsedrive}. Current geometric-aware retrieval methods operate primarily through three interconnected paradigms: Depth-driven approaches (e.g., ImOV3D~\cite{yang2024imov3d}), which reconstruct pseudo-point clouds from monocular RGB images but depend heavily on depth accuracy; direct 3D alignment, which leverages pre-scanned models for hierarchical feature matching~\cite{zhou2019dual}; and cross-modal fusion, which aligns semantics with 3D geometry via contrastive learning~\cite{hegde2023clip}, typically requiring synthetic point clouds. In parallel, 2D-centric strategies, including local–global descriptor fusion~\cite{yang2021dolg} and attention-based keypoint extraction~\cite{sarlin2021back}, remain spatially limited. Critically, most approaches rely on explicit 3D inputs—depth sensors~\cite{munaro20143d,danielczuk2019mechanical}, point clouds~\cite{wiesmann2022retriever}, or pre-rendered models~\cite{ausserlechner2024zs6d}—introducing deployment challenges in RGB-only settings and computational bottlenecks for real-time robotics.

Recent 3D foundation models such as the Visual Geometry Grounded Transformer (VGGT)~\cite{wang2025vggt} have demonstrated the ability to infer 3D geometry from multi-view 2D images using large-scale spatial priors. Leveraging VGGT-derived geometry, RoboEye enhances identification robustness in large-scale benchmarks while avoiding explicit 3D inputs.
\vspace{-5mm}

\begin{figure*}
    \centering
    \includegraphics[width=1\linewidth]{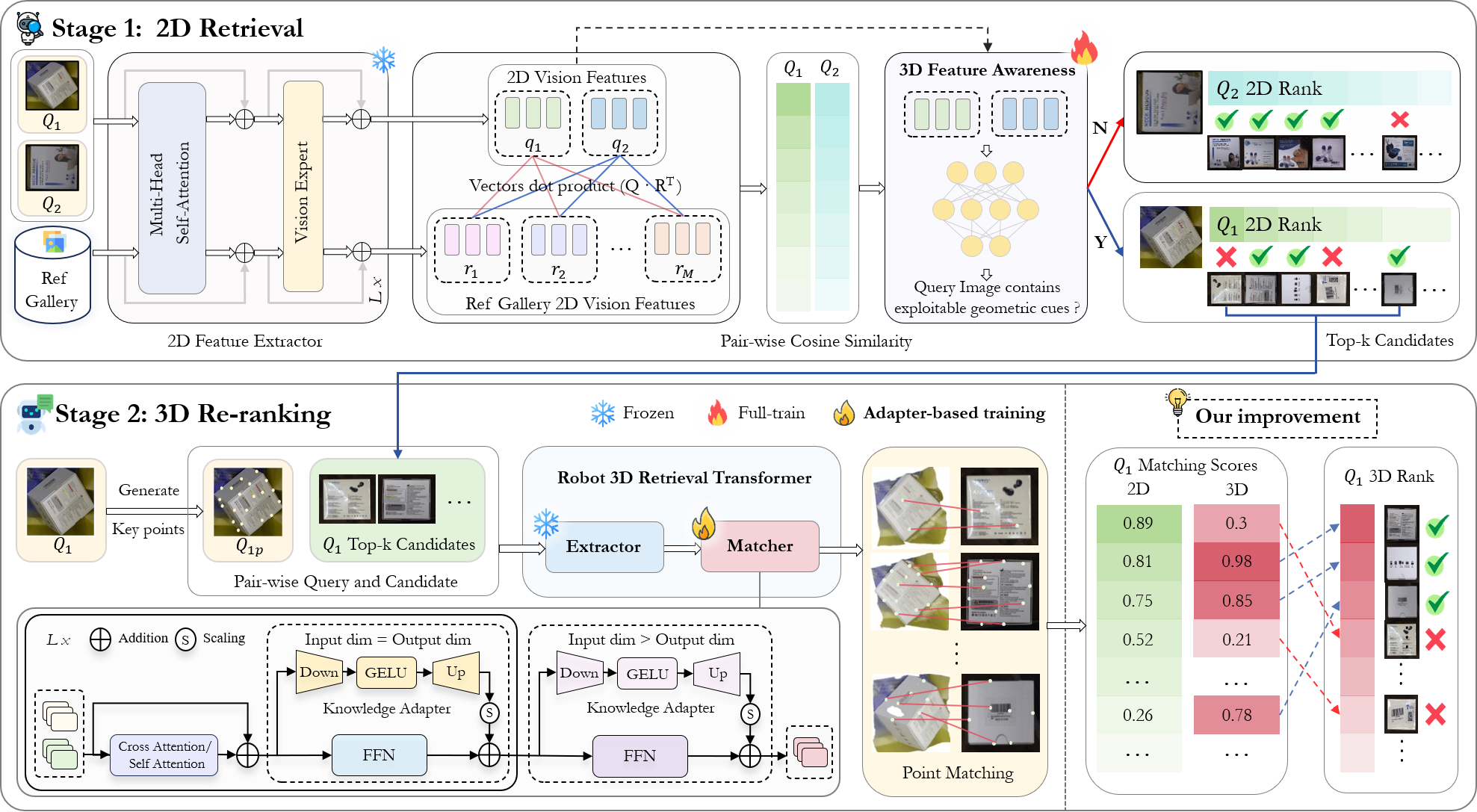}
    \caption{Overall architecture of \textit{RoboEye}. The framework follows a two-stage 2D$\rightarrow$3D retrieval pipeline: (i) 2D semantic retrieval with a feature extractor and a 3D-feature-awareness module that determines the need for geometric verification; (ii) 3D re-ranking using a transformer with a feature extractor and a keypoint-based matcher to compute confidence-driven correspondences. An adapter-based training scheme ensures efficient adaptation to warehouse-specific conditions.}
    \label{fig:placeholder}
    \vspace{-5mm}
\end{figure*}

\section{Method}

\subsection{Overview} 
As shown in Figure~\ref{fig:placeholder}, the proposed \textit{RoboEye} implements a two-stage warehouse object identification framework, dynamically augmenting 2D features with 3D cues. The first stage extracts discriminative 2D features to produce high-confidence candidates, with a lightweight 3D-feature-awareness module trained on an MRR-driven 3D-awareness objective deciding whether geometric reasoning is possible and necessary. If invoked, the second stage applies a robot 3D retrieval transformer comprising a 3D feature extractor and a keypoint-based matcher, which replaces cosine similarity with confidence-based keypoint correspondences for robust re-ranking under viewpoint shifts, occlusion, and packaging variation. To adapt these modules to warehouse conditions, we employ an efficient adapter-based training strategy, achieving strong performance under limited computational resources.


Subsequent sections first present the specific configurations for the Object ID task, followed by a detailed description of each stage of our framework and training methods.

\subsection{Task Definition} 

Given the inherent inefficiency and inaccuracy of classifying a vast number of categories by a dense layer, we formalize the Object ID task in warehouse operations as an image retrieval problem, where query images consist of segmented object patches acquired via instance segmentation and the reference gallery consists of predefined catalog images, each corresponding to a known object. In practice, query images appear in two configurations: (1) single-view setting, typical of pre-pick scenarios, where the system must identify an object from the cluttered and often occluded view inside the source container;
(2) multi-view setting, which arises when both pre-pick and post-pick images of the same object are available, offering complementary viewpoints that increase the likelihood of capturing reliable geometric cues. The core challenge lies in robustly matching query-gallery pairs despite significant geometric variations.

\subsection{Stage One: 2D Retrieval}

The initial stage involves an effective 2D-only feature retrieval process, where we generate a preliminary candidate ranking based on the cosine similarity of global appearance descriptors, followed by a 3D-feature-awareness module that adaptively determines the possibility and necessity of the subsequent 3D re-ranking stage.

\subsubsection{2D Feature Extractor}
The crucial 2D features serve as foundational anchors for our framework, enabling rapid candidate screening while preserving essential appearance cues that guide subsequent 3D geometric re-ranking. To meet the need for robust Object ID, we employ the pre-trained large model BEiT-3~\cite{wang2023image} with an architecture featuring the multiway transformer design and strong performance across different visual tasks. Each multiway transformer block comprises a shared self-attention module and a set of modality experts, optimized for distinct input modalities. In our implementation, we streamline this architecture for visual-only inputs by retaining solely the vision-specific expert, reducing parameter overhead, as shown in the upper left of Figure~\ref{fig:placeholder}.

To address varying levels of detail in Object ID, we consider two input scenarios: given a sequence of input images in the single-view setting $\left\{ Q_{i} \right\}_{i=1}^{N}$, or multi-view setting $\left\{ Q_{i}^{0}, Q_{i}^{1}, Q_{i}^{2} \right\}_{i=1}^{N}$, the feature extractor encodes these inputs into feature vectors. From the token dimension, we select the [CLS] token $\left\{q_{i} \right\}_{i=1}^{N} \in \mathbb{R}^{N \times d_{2D}}$ or $\left\{ q_{i}^{0}, q_{i}^{1}, q_{i}^{2} \right\}_{i=1}^{N} \in \mathbb{R}^{N \times \left\lfloor 3 \times d_{2D} \right\rfloor}$ as the global representation of each query, serving as a compact summary of its overall semantics. In the multi-view setting, concatenated [CLS] tokens proceed through an MLP to map the feature dimension, providing a fused representation. All reference images $\left\{ R_{i} \right\}_{i=1}^{M}$ undergo the same encoding pipeline, yielding [CLS] tokens $\left\{ r_{i} \right\}_{i=1}^{M} \in \mathbb{R}^{M \times d_{2D}}$. The similarity between each query–reference pair is computed by dot product of [CLS] tokens, yielding the initial 2D retrieval ranking. 

Building on this encoding pipeline, we further adapt the extractor to the warehouse environment by training it independently based on the methodology of RoboLLM~\cite{long2024robollm} using a contrastive loss to align the features for matched query–reference pairs.


\subsubsection{3D-feature-awareness Module}
Although 2D-only ranking demonstrates decent performance, challenging warehouse ID cases still require complementary 3D geometric cues, as explained in the introduction section. In exploring how to exploit 3D features, we identified three key issues. First, certain queries inherently lack sufficient geometric cues, as demonstrated in the upper part of Figure~\ref{fig:demo}. Second, indiscriminate 3D re-ranking can degrade accuracy because mismatches between 2D and 3D signals introduce noise that outweighs geometric benefits (see Section~\ref{sec:3dfeat}). Third, 3D re-ranking incurs substantial computational overhead. 

To address these issues, we propose a 3D-feature-awareness module that, given the initial 2D features, (1) assesses whether reliable 3D features can be extracted and (2) decides if 3D re-ranking is warranted. By skipping unnecessary 3D processing, the module reduces the second-stage inference cost and strikes a balance between accuracy and efficiency. Further details on the architectures and the training methodology of this module are presented in the following section.

\subsubsection{MRR-driven 3D-awareness Training}
We train the 3D-feature-awareness module to cooperate with 2D feature extractor using MRR-driven 3D-awareness Training (M3AT). The module is deliberately kept lightweight, consisting of only a few dense layers. While its size could be increased to achieve higher performance, doing so would also introduce greater computational cost and inference time. Its novelty and effectiveness arise from M3AT, a supervision scheme that identifies when 3D re-ranking provides tangible benefits and trains the module to detect implicit geometric cues within 2D features.


Concretely, for each query we compute the Mean Reciprocal Rank (MRR) of the initial 2D ranking and the MRR after applying 3D re-ranking. A query is labeled positive if its MRR improves after 3D re-ranking, and negative otherwise. To address the label imbalance, we optimize a class-weighted cross-entropy loss with class-specific weights \(w\), assigning a larger weight to positive examples.

The module maps a 2D feature \(q_i \in \mathbb{R}^{1 \times d_{2D}}\) to class logits via a two-layer projection:
\begin{align}
    h_i &= \operatorname{GELU}\big(\operatorname{LN}(q_i)\mathbf{W}_{\mathrm{hidden}}\big),\\
    \tilde{q}_i &= h_i \mathbf{W}_{\mathrm{class}} \in \mathbb{R}^2,
\end{align}
where \(\mathbf{W}_{\mathrm{hidden}}\in\mathbb{R}^{d_{2D}\times\tilde{d}}\) and \(\mathbf{W}_{\mathrm{class}}\in\mathbb{R}^{\tilde{d}\times 2}\). Here \(\operatorname{LN}\) denotes LayerNorm. The training objective is the weighted cross-entropy:
\[
\mathcal{L}_{\text{M3AT}} = \operatorname{CE}_{\!w}\big(\operatorname{softmax}(\tilde{q}_i),\, y_i\big),
\]
where \(y_i\) is the MRR-derived binary label and \(\operatorname{CE}_{\!w}\) applies weights \(w\) to the classes.

At inference, the module predicts whether to enable 3D re-ranking. If a query is predicted positive, we select the top-$K$ reference candidates from the initial 2D retrieval ranking and refine them using the 3D pipeline; otherwise the initial 2D ranking is returned. M3AT therefore concentrates the contribution on the training signal—learning when geometric reasoning is truly beneficial—while keeping the awareness module computationally trivial.



\subsection{Stage Two: 3D Geometric Re-ranking}

The lack of invariance in 2D appearance cues under warehouse settings amplifies discrepancies between query and reference images (shown in the lower part of Figure~\ref{fig:demo}). To address this, we introduce a robot 3D retrieval transformer, consisting of a 3D feature extractor and a 3D keypoint-based retrieval matcher (illustrated in the lower-left part of Figure~\ref{fig:placeholder}). The extractor encodes 3D cues into compact geometry-aware representations, while the matcher establishes dense keypoint correspondences between query and reference images, replacing conventional cosine similarity with a more robust confidence-based measure. Crucially, this module operates directly on 2D query-reference image pairs, avoiding the need for explicit 3D inputs such as point clouds or depth maps and augmenting 2D representations with a robust geometric verification.

\subsubsection{3D Feature Extractor}
We take the aggregator component in VGGT~\cite{wang2025vggt} as the 3D feature extractor in our framework. It employs a transformer architecture with alternating frame-wise and global self-attention layers. This design captures both intra-view spatial structure and cross-view geometric relations, yielding compact 3D-aware feature representations. These representations form the foundation for our proposed 3D point-based matching method.

\subsubsection{3D Keypoint-based Retrieval Matcher}
Once 3D dense features are extracted, the key challenge is how to leverage them effectively for retrieval. We observe that directly applying cosine similarity to 3D features provides poor discrimination (see Section~\ref{sec:3dfeat}). We argue it is because global 3D representations introduce additional noise for matching, whereas local geometric correspondences—conditioned on viewpoint and pose—are more discriminative than global fine-grained similarity. To address this limitation, we aim to design a 3D keypoint-based retrieval matcher that replaces cosine similarity with correspondence-driven scoring, providing more reliable compensation for 2D appearance-based ranking.


The track head of VGGT~\cite{wang2025vggt} partially meets our purpose and integrates well with the 3D feature extractor by leveraging large-scale end-to-end pretraining. The matcher is a transformer that processes a grid of point-match tokens, each initialized with appearance features and enriched with correlation cues for geometric alignment. Tokens encode position, visibility, and confidence, and are iteratively refined through interleaved self- and cross-attention, yielding dense correspondences across views.

However, the original track head was designed for point tracking, focusing primarily on finding correspondences across images. In contrast, our task requires a more sophisticated similarity measurement between query–reference pairs to support re-ranking and category identification. To this end, we adapt its mechanism to not only establish dense correspondences but also generate confidence scores as similarity estimates. We redesign it as a retrieval matcher that evaluates geometric consistency between candidate pairs, aggregating these scores into a ranking criterion that replaces conventional cosine similarity and yields more robust results.


Therefore, direct reuse of the track head in the VGGT model is ineffective due to both the shift in task objectives and the distribution gap between warehouse-specific datasets and VGGT’s pretraining data. To address the change in matcher's functionality, we introduce a 3D keypoint-based matching mechanism. To mitigate the distribution gap while ensuring practical deployment, we further develop an adapter-based domain adaptation strategy. The following sections provide detailed descriptions of these components.

\textbf{3D Keypoint-based Matching Mechanism.}
The top-$K$ query–reference image pairs $\left\{ Q_{i}, R_{i} \right\}_{i=1}^{K}$ obtained from the 2D semantic ranking serve as input to the re-ranking stage. 

The proposed matching mechanism is based on sparse keypoint matching rather than whole-image comparison. In the following, we describe the procedure for a single pair, which is applied independently to all candidates. The 3D feature extractor first processes the image pair through alternating frame-wise and global self-attention layers, producing a set of 3D tokens $T^{3D}$. Subsequently, the matcher takes the 3D tokens $T^{3D}$ to predict dense geometric tracking features ${F_{q}, F_{r}}$. Keypoints $\left( x_{j}, y_{j} \right)_{j=1}^{S}$ are first detected in each query image $Q_{i}$ using SIFT~\cite{lowe2004distinctive}, where $S$ denotes the number of keypoints. For each keypoint, its feature representation is extracted by bilinear sampling from the query feature map $F_{q}$ and then correlated with the feature map of reference candidate $F_{r}$. These correlation maps are processed through attention layers to predict matched keypoints $\left( \hat{x}_{j}, \hat{y}_{j} \right)_{j=1}^{S}$ together with confidence scores $\left\{ C_{j} \right\}_{j=1}^{S}$. Formally, the final similarity score $\tilde{C}$ for a candidate is computed as the mean confidence over all keypoints:
\begin{equation}
\tilde{C} = \frac{1}{S} \sum_{j=1}^{S} \left\{ C_{j} \right\}_{j=1}^{S}
\end{equation}
In the multi-view scenario, each query image independently produces a score $\tilde{C}$ for the same candidate, and the final re-ranking score is obtained by averaging across views. The sorted final scores represent the rank of the 3D geometric similarity between each query image and reference candidate pairs.


\textbf{Adapter-based Domain Shifting.} To bridge the domain gap between VGGT pretraining and warehouse operation conditions, the most direct approach would be to fully re-train VGGT with our proposed matching mechanism. However, this is computationally prohibitive, as the original setup requires 64 A100 GPUs for nine days~\cite{wang2025vggt}. To provide a practical alternative, we design an adapter-based training strategy: rather than updating the entire network, we freeze the 3D feature extractor and confine training to the matcher, enhanced with lightweight knowledge adapters~\cite{houlsby2019parameter}.

For the training objective, we construct samples by selecting the highest-ranked correct candidate for each query as a positive and the three top-ranked incorrect candidates as negatives. The matcher is optimized with a cross-entropy loss, which maximizes the scores of positives while suppressing those of negatives, enhancing the discriminative power of matcher for warehouse-specific scenarios.

As for the architecture, we use an adapter module to replace the feedforward network in a transformer block with a dual-path structure: a primary branch that retains the original pre-trained network, and a parallel, trainable branch dedicated to domain-specific adaptation. For parameter efficiency, this new branch employs a bottleneck architecture, featuring a down-projection and an up-projection layer. The transformation performed by the adaptive branch on an input feature $x_{i}^{\prime}$ to produce the new feature is formally expressed as:
\begin{equation}
\tilde{x_{i}} = \operatorname{GELU}(\operatorname{LN}(x_{i}^{\prime}) \cdot \mathbf{W}_{\text{down}}) \cdot \mathbf{W}_{\text{up}})
\end{equation}
In this formulation, $\mathbf{W}_{\text{down}} \in \mathbb{R}^{d \times \tilde{d}}$ and $\mathbf{W}_{\text{up}} \in \mathbb{R}^{\tilde{d} \times d^{\prime}}$ represent the down-projection and up-projection layers. LN stands for LayerNorm. $d^{\prime}$ can be set to match either the input dimension $d$ or the final output dimension of the transformer block and $\tilde{d}$ corresponds to the latent dimension of the bottleneck, satisfying $\tilde{d} \ll d$. The output of this bottleneck module is integrated with the original FFN path via a scaled residual connection modulated by a factor $\alpha$. Subsequently, the combined features from both paths, $\tilde{x_{i}}$ and $x_{i}^{\prime}$, are fused with the initial input $x_{i}$ through a final residual connection:
\begin{equation}
x_{i} = FFN\left(LN\left( x_{i}^{\prime} \right) \right) + \alpha \cdot \tilde{x}_{i} + x_{i}^{\prime}
\end{equation}

\section{EXPERIMENTS}

\subsection{Experimental Setup}

\textbf{Dataset and Configuration.}
The experiments use the ARMBench dataset~\cite{mitash2023armbench} provided by Amazon, a large-scale benchmark containing over 190,000 unique items and reflecting realistic warehouse conditions. We consider two query settings: (1) single-view: one pre-pick image per item; (2) multi-view: the pre-pick image augmented by two post-pick images. We evaluate under two reference-gallery scenarios: (a) container gallery, where entries correspond to containers holding multiple objects; and (b) global gallery, containing all unique objects across the dataset.


\textbf{Training Details.} The results reported for the BEiT-3 Large model are based on a training methodology consistent with that of the 2D feature extractor in RoboEye. Both the 2D feature extractor and the corresponding components of the robot 3D retrieval transformer are initialized with their pre-trained weights, while knowledge adapters and the 3D-feature-awareness module are randomly initialized. The hidden dimension of the 3D-feature-awareness module is set to 64, equal to the hidden dimension of the knowledge adapter. 

Following previous works~\cite{long2024robollm,mitash2023armbench,gouda2024learning}, performance is measured with Recall@\(k\). The cross-entropy loss for MRR-driven 3D-awareness training uses class-specific weights \(w\) with a 4:1 ratio. During Object Identification experiments, we set the number of candidates for re-ranking to 16 and sample 20 keypoints.

\begin{table}[t]
\centering
\resizebox{0.48\textwidth}{!}{\begin{tabular}{lc|c|c|c|c|c|c}
\toprule
\multirow{2}{*}{Model} & \multirow{2}{*}{Ref Set} & \multicolumn{2}{c}{Recall@1} & \multicolumn{2}{c}{Recall@2} & \multicolumn{2}{c}{Recall@3} \\
\cmidrule(lr){3-4} \cmidrule(lr){5-6} \cmidrule(lr){7-8}
 & & N=1 & N=3 & N=1 & N=3 & N=1 & N=3 \\
\midrule
ResNet50-RMAC & Container & 71.7 & 72.2 & 81.9 & 82.9 & 87.2 & 88.2 \\
DINO-ViT-S & Container & 77.2 & 79.5 & 87.3 & 89.4 & 91.6 & 93.5 \\
BEiT-3-Base* & Container & 83.7 & 84.5 & 83.8 & N/A & 84.5 & N/A \\
BEiT-3-Large & Container & 96.9 & 99.0 & 97.0 & 99.1 & 97.2 & 99.2 \\
RoboLLM & Container & 97.8 & 98.0 & 97.9 & 98.1 & 98.0 & 98.2 \\
RoboEye & Container & \textbf{98.2} & \textbf{99.4} & \textbf{98.3} & \textbf{99.5} & \textbf{98.4} & \textbf{99.6} \\
\midrule
BEiT-3-Large & Global & 62.4 & 35.1 & 69.3 & 46.7 & 72.4 & 53.3 \\
RoboLLM & Global & 74.6 & 78.2 & 82.6 & 85.7 & 85.3 & 89.1 \\
RoboEye & Global & \textbf{79.4} & \textbf{85.3} & \textbf{84.4} & \textbf{91.1} & \textbf{86.3} & \textbf{93.5} \\
\bottomrule
\end{tabular}}
\caption{Results on the ID task at varying Recall@k. N=1 denotes the single-view setting, while 
N=3 denotes the multi-view setting. * denotes no ARMBench training. ``Ref Set'' indicates whether the reference images are drawn from the container gallery or the global gallery.}
\vspace{-4mm}
\label{tab:object_identification_results}
\end{table}

\subsection{Object Identification in Single-view Setting}
We begin our evaluation with the single-view setting, the most common yet constrained condition. Due to the large-scale catalog and the presence of viewpoint shifts, occlusions, and packaging variations, the proposed framework must correctly identify objects under restricted viewpoint coverage and with limited 3D geometric information.

As summarized in Table~\ref{tab:object_identification_results}, RoboEye consistently outperforms previous state-of-the-art methods. In the most challenging case—the global gallery setting—RoboEye achieves a 4.8\% improvement at Recall@1. Under the container gallery setting, RoboEye raises Recall@1 from 97.8\% to 98.2\% (+0.4\%) compared to RoboLLM, despite the strong, near-saturated baseline. Note that the larger 2D baseline, BEiT-3 Large, suffers a sharp performance drop to only 62.4\% on Recall@1 as the catalog expands, whereas RoboEye remains robust with 79.4\%. This suggests that merely scaling up model size does not necessarily improve performance for this task. In contrast, RoboEye effectively counteracts degradation from viewpoint shifts, occlusions, packaging variations, and large-scale catalog growth, delivering consistent gains even with minimal inputs.

\subsection{Object Identification in Multi-view Setting}
This section evaluates the performance of RoboEye in a multi-view setting, where the framework leverages cross-view feature fusion and 3D geometric cues to enhance consistency and accuracy in object identification.

The quantitative comparison is shown in Table~\ref{tab:object_identification_results}. RoboEye achieves larger absolute gains in this setting compared to the single-view setting, as the additional inputs reinforce its core mechanism. When retrieving from the container gallery, Recall@1 improves from 98.0\% to 99.4\% (+1.4\%) compared to RoboLLM. In the global gallery retrieval scenario, Recall@1 of RoboEye increases by 7.1\%, establishing a clear margin over the best prior result. These findings demonstrate that RoboEye generalizes beyond single-query conditions and becomes more effective with multiple queries, further highlighting its robustness in practical warehouse environments. The framework not only maintains consistency across views but also fully exploits multi-view redundancy through feature fusion and cross-image 3D geometric reasoning.


\begin{table}[t]
\centering
\resizebox{0.49\textwidth}{!}{
\begin{tabular}{lcccccc}
\toprule
 2D-FE & 3D-FE & 3D-KMNT & 3D-KMT & 3D-FAM & Recall@1 \\
\midrule
 $\cmark$ & $\xmark$ & $\xmark$ & $\xmark$ & $\xmark$  & 70.5 \\
 $\xmark$ & $\cmark$ & $\xmark$ & $\xmark$ & $\xmark$  & 2.0 \\
 $\cmark$ & $\cmark$ & $\xmark$ & $\xmark$ & $\xmark$  & 18.2 \\
 $\cmark$ & $\cmark$ & $\cmark$ & $\xmark$ & $\xmark$  & 64.4 \\
 $\cmark$ & $\cmark$ & $\cmark$ & $\cmark$ & $\xmark$  & 68.2 \\
 $\cmark$ & $\cmark$ & $\cmark$ & $\cmark$ & $\cmark$  & \textbf{79.4} \\
\bottomrule
\end{tabular}
}
\caption{Ablation study of RoboEye components: 2D-FE (2D feature extractor), 3D-FE (3D feature extractor), 3D-KMNT (3D keypoint-based retrieval matcher without warehouse-specific training), 3D-KMT (3D keypoint-based retrieval matcher with warehouse-specific training), and 3D-FAM (3D-feature-awareness module).}
\vspace{-4mm}
\label{tab:ablation_results}
\end{table}

\section{Analysis}

\subsection{Analysis Configuration}
Unless otherwise stated, all analyses are conducted under the single-view setting with retrieval from the global gallery, as this represents the most common and challenging scenario. The 2D feature extractor is implemented in two variants: BEiT-3 Base and BEiT-3 Large, with the latter used only in comparison experiments to separate the effects of increased parameters from the architectural contributions of RoboEye.

\subsection{Naive 3D Fusion vs. Selective Geometric Re-ranking}
\label{sec:3dfeat}

In this section, we evaluate the contributions of RoboEye’s key components and outline the exploration process that led to its design. We first experimented with using only 3D geometric features, but this approach achieves poor results. We then tried a naive two-stage pipeline, using 2D features for initial retrieval and 3D features for re-ranking; however, this still underperformed compared to using 2D features alone. These findings motivated us to develop a dedicated matching mechanism and training strategy for 3D geometric retrieval, rather than relying solely on cosine similarity for measuring feature similarity. This progression ultimately resulted in the RoboEye framework proposed in this paper.


Using the 2D feature extractor alone, we achieve a Recall@1 of 70.5\% (first line in Table~\ref{tab:ablation_results}). Although we use the same backbone and training procedure as RoboLLM~\cite{long2024robollm}, RoboLLM reports 74.6\% under this setting. We attribute this gap to hardware limitations (smaller batch sizes), since contrastive learning typically benefits from larger batches. Despite our 2D model being 4.1\% weaker initially, our full framework (RoboEye) still outperforms RoboLLM by 4.8\%, demonstrating its effectiveness.

Next, we evaluate the performance using only the 3D geometric feature extractor. This experiment yields a very low Recall@1 of 2.0\% (second line in Table~\ref{tab:ablation_results}), indicating that directly using cosine similarity on 3D geometry as the similarity measurement is insufficient for retrieval. This poor performance motivates designing a specialized matching mechanism and a matcher model to better leverage 3D features.

We also tried a naive two-stage pipeline, in which 2D features perform the initial ranking followed by 3D feature re-ranking. However, this simple fusion only marginally improves performance: Recall@1 rises from 2.0\% to 18.2\% (third line in Table~\ref{tab:ablation_results}), which remains far below the 2D-only baseline of 70.5\%. This suggests that straightforward 3D-based re-ranking introduces significant noise, undermining the discriminative power of the 2D features.

To fully exploit multi-view 3D information, we evaluate the 3D keypoint-based retrieval matcher with two variants. The first has no warehouse-specific adaptation (3D-KMNT); the second includes adapter-based training on warehouse data (3D-KMT). Using 3D-KMNT raises Recall@1 to 64.4\% (fourth line in Table~\ref{tab:ablation_results}), yet this is still below the 2D-only baseline, highlighting a domain gap in applying generic geometric reasoning to warehouse images. With warehouse-adapted training (3D-KMT), Recall@1 improves to 68.2\% (+3.8\%, fifth line in Table~\ref{tab:ablation_results}), confirming that lightweight domain adaptation helps. Still, this is below the 2D baseline, indicating that geometry alone cannot match the state-of-the-art ID performance.

Finally, we introduce a 3D-feature-awareness module to dynamically combine 3D and 2D features. This module assesses whether reliable 3D information can be extracted from a given query and whether it should be applied. By selectively enabling 3D-based re-ranking only when beneficial, we avoid performance degradation from noisy 3D data and unnecessary computation. As shown in the last line of Table~\ref{tab:ablation_results}, adding the 3D-feature-awareness module yields the best result, increasing Recall@1 from 68.2\% to 79.4\% (+11.2\%). Integrating all components together achieves the highest performance, demonstrating that domain-adapted 3D reasoning with selective activation is critical for accurate and efficient large-scale warehouse identification.

\subsection{How Does RoboEye Reduces Inference Latency?}

\begin{table}[t]
\centering
\resizebox{0.49\textwidth}{!}{
\begin{tabular}{lccccccc}
\toprule
Model & 2D-FE(Base) & 2D-FE(Large) & 3D-T & 3D-FAM & Time & Recall@1 \\
\midrule
RoboEye & $\cmark$ & $\xmark$ & $\xmark$ & $\xmark$ & \textbf{0.028} & 70.5 \\
RoboEye & $\cmark$ & $\xmark$ & $\cmark$ & $\xmark$ & 0.547 & 64.4 \\
RoboEye & $\cmark$ & $\xmark$ & $\cmark$ & $\cmark$ & 0.071 & \textbf{76.8} \\
RoboEye & $\xmark$ & $\cmark$ & $\xmark$ & $\xmark$ & 0.068 & 62.4 \\
\bottomrule
\end{tabular}
}
\caption{Average per-sample inference time (seconds) under different configurations. ``2D-FE(Base)'' and ``2D-FE(Large)'' denote the base and large variants of the 2D feature extractor, respectively. ``3D-T'' indicates the robot 3D retrieval transformer, and ``3D-FAM'' denotes the 3D-feature-awareness module.}
\label{tab:time_results}
\vspace{-3mm}
\end{table}

To understand the trade-off between recognition accuracy and computational efficiency, we measure the average per-sample inference latency and Recall@1 across different configurations, as summarized in Table~\ref{tab:time_results}. Results are reported with candidate pool size set to 4, measured on a single NVIDIA 5090 GPU with 32GB memory. The first baseline uses only the BEiT-3 Base model and achieves the fastest runtime with moderate accuracy, providing efficient candidate retrieval but without 3D geometric verification. Incorporating the robot 3D retrieval transformer without any awareness control (see second line in Table~\ref{tab:time_results}) increases latency from 0.028 to 0.547 seconds while even reducing Recall@1 by 6.1\%, reflecting the computational burden of unconditional 3D re-ranking without performance gain. With the 3D-feature-awareness module (third line in Table~\ref{tab:time_results}), Recall@1 improves by 6.3\% while latency remains low at 0.071 seconds—nearly an order of magnitude faster than naive 3D re-ranking. The latency is also comparable to the large 2D feature extractor (0.068 seconds, last line in Table~\ref{tab:time_results}), yet achieves 14.4\% higher Recall@1 and retains the advantages of geometric verification. 

These results highlight the pivotal role of the 3D-feature-awareness module in balancing efficiency and robustness. By activating 3D reasoning only when necessary, it preserves near-2D runtime while retaining the benefits of selective geometric verification, making the approach practical for large-scale warehouse deployments where both latency and reliability are critical.

\subsection{Does a Larger Model Always Help?}

\begin{table}[t]
\centering
\resizebox{0.49\textwidth}{!}{
\begin{tabular}{lcccc}
\toprule
Model & Trained Parameters & Recall@1 & Recall@2 & Recall@3 \\
\midrule
2D-FE(BEiT-3 Base) & 221.0M & 70.5 & 78.2 & 81.7 \\
2D-FE(BEiT-3 Large) & 672.7M & 62.4 & 69.3 & 72.4 \\
RoboEye & 222.3M & \textbf{79.4} & \textbf{84.4} & \textbf{86.3} \\

\bottomrule
\end{tabular}
}
\caption{Comparison of retrieval performance and trained parameter scales between RoboEye and 2D feature extractors of different model sizes.}
\label{tab:compare_results}
\vspace{-4mm}
\end{table}

In this section, we investigate whether simply increasing the size of 2D feature extractors can obviate the need for exploiting 3D geometric features. To this end, we perform a controlled comparison between the full RoboEye pipeline and two variants of the 2D feature extractor with different parameter sizes.

As shown in Table~\ref{tab:compare_results}, scaling the BEiT-3 extractor from 221.0M to 672.7M parameters not only increases training cost but also leads to degraded performance, with Recall@1 dropping by 8.1\%. In contrast, RoboEye achieves an improvement of 17.0\% in Recall@1 over the BEiT-3 Large model while using only 222.3M trained parameters—essentially the same scale as BEiT-3 Base and merely one-third the size of BEiT-3 Large—yet delivering a clear performance gain.


This comparison shows that enlarging the 2D backbone alone is insufficient to overcome warehouse-specific challenges such as occlusion, clutter, packaging variation, and inter-class visual similarity. By contrast, RoboEye leverages complementary geometric reasoning, enabled by its robot 3D retrieval transformer and 3D awareness-guided re-ranking, to provide robust improvements at a comparable parameter scale. These results underscore the necessity of incorporating 3D-aware components rather than relying solely on larger 2D models for effective large-scale warehouse ID.

\subsection{Impacts of 2D Candidate Pool Size}


\begin{table}[t]
\centering
\resizebox{0.48\textwidth}{!}{\begin{tabular}{lcccc}
\toprule
Model & Candidates & Recall@1 & Recall@2 & Recall@3 \\
\midrule
RoboEye & 4 & 76.8 & 81.0 & 82.7 \\
RoboEye & 8 & 78.5 & 83.1 & 84.9 \\
RoboEye & \textbf{16} & \textbf{79.4} & \textbf{84.4} & \textbf{86.3} \\
RoboEye & 32 & 78.8 & 83.8 & 85.8 \\
RoboEye & 64 & 78.5 & 83.3 & 85.3 \\
\bottomrule
\end{tabular}}
\caption{ID performance of RoboEye with varying candidate pool sizes ($K$) after applying the full two-stage pipeline.}
\label{tab:candidates_results_1}
\vspace{-4mm}
\end{table}
In this section, we evaluate the impact of the candidate-pool size $K$, introduced by RoboEye's two-stage ranking design, on final performance.

We first evaluate the recall of the 2D feature extractor alone at varying cutoff thresholds $K$. This measurement defines the upper bound of stage two performance, as 3D re-ranking is based on the 2D ranking result and any candidate not retrieved in stage one cannot be recovered by subsequent re-ranking. Specifically, recall rises from 84.7\% at a pool size of 4 to 87.4\%, 90.5\%, 92.6\%, and 94.3\% as the pool expands to 8, 16, 32, and 64, respectively. Expanding the pool size from 4 to 16 yields a 5.8\% gain, whereas enlarging the pool from 16 to 64 (a fourfold increase) brings only a modest 3.8\% improvement. While larger candidate pools provide slightly higher coverage of the ground-truth object, they also impose substantially greater computational cost during 3D re-ranking. Conversely, smaller pools risk excluding the correct object, thereby capping overall performance.  

Beyond the upper-bound analysis, we further investigate how different candidate pool sizes affect the end-to-end performance of RoboEye. The results in Table~\ref{tab:candidates_results_1} reveal several important trends. Increasing the pool size from 4 to 16 raises Recall@1 by 2.6\%, yielding the best overall results. However, further enlarging the pool to 32 or 64 candidates does not yield additional gains; instead, performance slightly declines (Recall@1 drops by 0.6\% and 0.9\%, respectively). This counterintuitive trend suggests that including too many low-quality candidates introduces noise into the geometric verification stage, which can dilute discriminative signals and destabilize re-ranking. Most importantly, these results indicate that RoboEye is insensitive to the candidate-pool size 
$K$: values from 8 to 64 yield comparable performance. Taken together, we adopt $K=16$ as the default configuration, which strikes a balance between retrieval coverage, re-ranking effectiveness, and computational cost.


\section{CONCLUSIONS}

We presented RoboEye, a two-stage identification framework that augments 2D appearance features with selective 3D reasoning. By dynamically invoking 3D re-ranking only when beneficial, RoboEye achieves robust performance without explicit 3D inputs. Experiments on the Amazon ARMBench dataset show up to 7.1\% Recall@1 improvement over the previous state of the art, demonstrating that RoboEye effectively mitigates the challenges posed by large-scale catalogs, viewpoint and pose variations, occlusions, and packaging appearance changes, thereby establishing RoboEye as an efficient and scalable solution for reliable object identification in warehouse automation.
\vspace{3mm}

\addtolength{\textheight}{-12cm}   




\bibliographystyle{IEEEtran}
\bibliography{Mybib}

\end{document}